\pdfoutput=1
\documentclass[sigconf,nonacm]{acmart}

\usepackage{algorithm}
\usepackage{algpseudocode}
\usepackage{booktabs}
\usepackage{placeins}

\newcommand{\method}{\textsc{CohortMix-TS}}
\newcommand{\cset}{\mathcal{C}}
\newcommand{\aset}{\mathcal{A}}
\newcommand{\qset}{\mathcal{Q}}
\newcommand{\uset}{\mathcal{U}}

\AtBeginDocument{%
  }
\setcopyright{none}

\begin{document}

\title{Challenges and Solutions for Bandits in the Wild: Warm-Started Mixture Bandits for Cross-Cohort Slate Recommendation}

\author{%
Serafima Lebedeva\textsuperscript{1,2,$\ast$},\quad
Sumantrak Mukherjee\textsuperscript{1,$\ast$},\quad
Ali Arshad Sadal\textsuperscript{2},\quad
Ilias Ek\c{s}i\textsuperscript{2},\quad
Rahul Sharma\textsuperscript{1,2},\quad
Julia Mueller\textsuperscript{2},\quad
Theresa Dombrowski\textsuperscript{2},\quad
Jakob Karolus\textsuperscript{1,2},\quad
Viktor Bengs\textsuperscript{1},\quad
Eyke H\"ullermeier\textsuperscript{1,3},\quad
Sebastian Vollmer\textsuperscript{1,2}%
}
\authornote{\textsuperscript{1}\, German Research Center for Artificial Intelligence (DFKI), Kaiserslautern;\
\textsuperscript{2}\,RPTU, Kaiserslautern-Landau;\  \textsuperscript{3}\,LMU,  Munich.\ \textsuperscript{$\ast$}\,Equal contribution: serafima.lebedeva@dfki.de, sumantrak.mukherjee@dfki.de.}
\affiliation{\institution{~}\country{~}}

\renewcommand{\shortauthors}{ Lebedeva and Mukherjee et al.}

\begin{abstract}
Many recommender services repeatedly encounter cold-start cohorts, where new users arrive with little or no interaction history. This creates two challenges: learning user preferences quickly from limited feedback and sustaining useful recommendations when each user has a finite catalog that can become repetitive or depleted over time. We propose \method{}, a warm-started mixture bandit that learns latent user groups from earlier cohorts and uses available metadata to construct group-informed priors for new users. Starting from these fixed priors, the model personalizes independently as feedback from each user becomes available. Session slates combine Thompson sampling with diversity and inventory-depletion controls.
We evaluate \method{} through simulation, semi-synthetic experiments, and a 25-day randomized in-the-wild deployment with 713 registered participants in a Campus Games quiz application. Our evaluations show that cross-cohort transfer improves early recommendation quality and user-level regret, while inventory-aware slate construction helps prevent premature exhaustion of preferred items. In the field deployment, treatment users also showed a larger early-to-late change in correctness than users receiving random recommendations. Together, these results show how warm-start transfer and inventory-aware recommendations can support personalization for short-lived, repeatedly cold-starting cohorts.
\end{abstract}


\maketitle
\hypersetup{
  pdfauthor={Serafima Lebedeva, Sumantrak Mukherjee,
    Ali Arshad Sadal, Ilias Ek\c{s}i, Rahul Sharma,
    Julia Mueller, Theresa Dombrowski, Jakob Karolus,
    Viktor Bengs, Eyke H\"ullermeier, Sebastian Vollmer}
}

\section{Introduction}
\label{sec:introduction}

Recommendation systems select items in video platforms \cite{davidsonYouTubeVideoRecommendation2010}, retail services \cite{gomez-uribeNetflixRecommenderSystem2016a}, news systems \cite{karimiNewsRecommenderSystems2018}, and mobile applications \cite{maqboolMobileRecLargeScale2023}; broader accounts describe recommender systems \cite{lior2011introduction} and their evolution \cite{smithTwoDecadesRecommender2017}. In several applications, users are recruited or activated in cohorts. Flash-sale periods \cite{li2022hml4rec} create short-lived recommendation tasks as available assortments change, while online-learning systems \cite{ma2023doris} repeatedly select activities for new learners \cite{choiKnowledgeTracing2020}. Campaign-style games and health interventions \cite{sweenRoleExergamingImproving2014} similarly place many participants on a common schedule. A \emph{cohort} is active for a short horizon, so the recommender must personalize early enough to make the displayed items useful to each user.

New cohort members commonly arrive without a current interaction history or a complete context for personalization. The operational setting also differs from an unconstrained ranking problem: items cannot usually be repeated for a user, a user's candidate pool is finite, and some users disengage before the campaign ends. Learning a user's preferences from feedback is therefore a sequential decision problem. Multi-armed bandits provide a standard framework for this problem \cite{barraza2020banditstutorial}, and surveys document their use in recommendation \cite{silva2022mabsurvey}. Thompson sampling \cite{agrawal2013tsregret} balances exploration of uncertain items with exploitation of items whose current reward estimates are high; information-directed sampling \cite{russo2016infoTS} provides a related Bayesian decision rule. In a short campaign, however, exploration that would pay off only after a long horizon can consume a material share of the available interactions. Moreover, a policy that repeatedly favors one item type can use its finite per-user inventory too early. Item similarity could support generalization across items, but sufficiently rich item metadata are not always available to define that similarity directly.

Collaborative filtering learns long-term user--item structure from earlier interactions, but a newly arriving user has not supplied the observations needed for an individual factor estimate. It also does not prescribe how to choose sequential slates or how much uncertainty to explore. Earlier cohorts nevertheless contain population-level preference structure that can provide a warm start for the next cohort. We use that structure as an uncertain, metadata-conditioned prior rather than as a fixed user assignment: users with the same metadata can still belong to different preference groups, and current feedback can separate them.

We address this setting with \method, a cross-cohort mixture bandit. It transfers historical item and user structure into soft priors, constructs feasible slates through an arm-level Thompson rule, and performs scheduled batch updates between sessions. The empirical study is central to the paper: we pair controlled simulations with a 25-day randomized in-the-wild deployment of the full policy in the Campus Games of the RPTU \footnote{ Code is available at \url{https://github.com/etowho/university-games}.}. Four questions organize this study: \textbf{RQ1} asks whether historical warm starts improve early decisions; \textbf{RQ2} asks whether soft memberships help when users with the same metadata have different preferences; \textbf{RQ3} asks whether diversity and depletion controls preserve useful items for later rounds; and \textbf{RQ4} asks whether the complete policy differs from random recommendation in the in-the-wild study.

This paper makes the following contributions:
\begin{itemize}
    \item We formulate cross-cohort slate recommendation with historical cohort data, cold-start users, delayed binary feedback, user-level no-repeat constraints, finite per-user inventories, and changing active-user sets.
    \item We introduce \method, which combines historical latent structure, soft metadata-conditioned priors, Thompson sampling, diversity and depletion controls, and scheduled approximate hierarchical updates.
    \item We evaluate the mechanisms in two parametric environments and a source-only 2024-to-2025 data-calibrated semi-synthetic environment with replication-based uncertainty estimates.
    \item We report a randomized 25-day in-the-wild deployment of the complete policy against random recommendation with 713 registered participants.
\end{itemize}

Section~\ref{sec:related} positions the work, Section~\ref{sec:problem} defines the setting, and Section~\ref{sec:method} describes the method. Sections~\ref{sec:experiments} and \ref{sec:results} give the evaluation design and results, respectively.

\section{Related Work}
\label{sec:related}

\paragraph{Cold-start recommendation and transfer.}
Collaborative filtering estimates low-dimensional user--item structure from observed interactions \cite{koren2009cf}; implicit-feedback variants model a related signal \cite{hu2022implicitmf}. Factorization bandits \cite{wangFactorization2017} and collaborative-filtering bandits \cite{liCollaborativeFiltering2016} combine such shared representations with sequential item selection. Matrix-factorization variants can represent multiple user tastes \cite{chen2017wemarec,chen2016mpma}. Long-running datasets, including MovieLens \cite{harperMovieLensDatasetsHistory2015} and MIND \cite{wuMINDLargescaleDataset2020}, provide the interaction traces on which such models rely. A new user faces the cold-start problem \cite{bobadilla2012coldstart}; an occasional visitor may likewise lack direct feedback at arrival \cite{aharon2015yahooanswers}. Cold-start methods therefore use side information, population structure, or transferred representations to initialize a model. Meta-bandits \cite{kvetonMetaTS2021}, hierarchical bandits \cite{hongHierarchicalBandits2022}, and warm-start bandits \cite{oetomoWarmStart2023} likewise transfer shared structure across related tasks. Dynamic collaborative filtering \cite{ishikawaDynamicCollaborativeFiltering2022}, empirical-Bayes multi-bandits \cite{jiangEmpiricalBayesianMultiBandit2025}, and empirical-Bayes meta-prior learning \cite{nabiMetaPrior2022} offer related approaches. \method{} instead uses historical factors to form item arms and metadata-conditioned priors, then revises arm beliefs during the short target cohort.

\paragraph{Online, contextual, and clustered bandit recommendation.}
Bandits formalize the choice between gathering information and choosing items with high current reward estimates. Contextual bandits use observed features \cite{liContextualBandit2010,agrawal2013contextualts}, while clustered variants share information through inferred groups \cite{yang2020clexb,mahadik2021distclub}. A contextual model restricted to categorical metadata learns pooled metadata--arm effects but does not maintain a user-specific latent state; we evaluate this representation with Metadata LinUCB and Metadata LinTS. Our scheduled updates are operational feedback boundaries, rather than adaptive batching \cite{kalkanliBatchedTS2021}. Field recommender studies \cite{garcinOfflineOnlineEvaluation2014,mcinerneyExploreExploitExplain2018} and large production decision systems \cite{mcmahanAdClickPrediction2013} have shown why deployed behavior can differ from offline metrics. Partial feedback complicates unbiased offline evaluation \cite{liUnbiasedOfflineEvaluation2011}; treatment-style recommendations complicate causal evaluation \cite{schnabelRecommendationsTreatments2016}; and evaluation surveys summarize these concerns \cite{zangerleEvaluatingRecommenderSystems2022}. This motivates both randomization and careful outcome definitions.

\paragraph{Constrained slate recommendation.}
Slate-aware rankers model interactions in a displayed list \cite{renSlateAware2023}, and slate-likelihood Thompson policies can diversify it \cite{eideDynamicSlate2022}. Diversified contextual combinatorial bandits use determinantal point processes for a richer relevance--diversity trade-off \cite{liuDC2B2020}. These approaches are complementary to our arm-level selector: a slate-aware re-ranker could operate on feasible within-arm candidates. Multi-item mobile recommendation also exhibits slate and position effects \cite{heContextualUserBrowsing2020}. Selecting a session slate changes future availability. Sleeping bandits motivate changing action sets \cite{kleinberg2010regret}; rotting bandits address repeated use \cite{levine2017rotting}; and non-stationary bandit methods offer related tools \cite{seznec2019rotting}. Our availability instead comes from user-level no-repeat constraints rather than a globally disappearing catalog. We use a penalty score rather than claim an exact constrained-bandit solution.

Together, these lines leave a practical gap: a policy must transfer uncertain cohort structure, adapt within a short campaign, and construct feasible multi-item sessions from a finite per-user catalog.

\section{Problem Formulation}
\label{sec:problem}

We write $[n]=\{1,\ldots,n\}$. Let $\qset$ denote the fixed item catalog, $T$ the number of rounds, and $K$ the number of items presented to each active user in a round. The catalog is the same in every round; a user's available subset changes only after items have been displayed to that user.

\subsection{Cohorts, information, and indexing}
\label{sec:notation}

In our setting, a set of users who join during the same time period forms a \emph{cohort}. Let $\uset$ be the users in the current cohort. Historical data $\mathcal{D}^{\mathrm{hist}}$ contain records from earlier cohorts of the items displayed to a user and the resulting binary outcome. Each new user $u\in\uset$ arrives with an observed metadata group $g_u$, but with no outcome from the current cohort.

We index rounds by $t\in[T]$, items by $q\in\qset$, and users by $u\in\uset$. Let $\uset_t\subseteq\uset$ be the users who are active in round $t$. The active set can shrink when users leave or do not complete a round.

\subsection{Cross-cohort slate environment}
\label{sec:environment}

For each $u\in\uset_t$, the learner presents a slate $S_{u,t}\subseteq\qset$ of $K$ distinct items. The outcome for item $q\in S_{u,t}$ is $y_{u,t,q}\in\{0,1\}$. Before round $t$, the items previously displayed to user $u$ are
\[
\qset^{\mathrm{seen}}_{u,t-1}=\bigcup_{\tau<t}S_{u,\tau}.
\]
The learner therefore selects $S_{u,t}\subseteq\qset\setminus\qset^{\mathrm{seen}}_{u,t-1}$. It receives the outcomes for a slate after the round, then uses the accumulated observations to form later slates. The final slate is a set: its display order is not part of the model.

\subsection{Objective and evaluation}
\label{sec:objective}

The purpose of personalization is to make useful choices for every user throughout the $T$ rounds. We therefore seek high reward in each round and evaluate the learner by cumulative reward over the cohort horizon:
\[
\max_{\pi}\;\mathbb{E}_{\pi}\!\left[\sum_{t=1}^{T}\sum_{u\in\uset_t}\sum_{q\in S_{u,t}}y_{u,t,q}\right].
\tag{1}
\]
The policy $\pi$ uses $g_u$, $\mathcal{D}^{\mathrm{hist}}$, and outcomes observed before round $t$ when selecting a feasible slate.

In the synthetic environments, the evaluator knows the generated reward probabilities and can report pseudo-regret as an additional measure of how efficiently the learner personalizes. Those probabilities are unavailable in the in-the-wild study, so we compare the change in observed performance for the personalized learner and the random control. For user $u$, the scalar early--late change is
\[
\Delta_u=C_u(W_u^{\mathrm{late}})-C_u(W_u^{\mathrm{early}}),
\tag{2}
\]
where $C_u(W)$ is that user's mean of daily correctness values over observed days in time window $W$. Users can also leave the study or complete only some rounds. Better personalization may improve continued participation, so we report participation and completion accounting as secondary field measures; cumulative reward remains the primary optimization metric.

\section{\method: Warm-Started Mixture Bandits}
\label{sec:method}

\subsection{Historical structure and cold-start priors}
\label{sec:priors}

Earlier cohorts commonly provide response records for the same finite catalog while a newly arriving user has no current-cohort outcomes. We use those records to group items into arms and to form priors that make initial personalization possible. Let $R\in[0,1]^{n_H\times|\qset|}$ be the sparse historical user--item response matrix, where $R_{h,q}$ is the mean binary outcome for historical user $h$ and item $q$. Let $\Omega\subseteq[n_H]\times\qset$ be the set of observed entries of $R$. We fit a rank-$d$ regularized matrix factorization, where $U\in\mathbb{R}^{n_H\times d}$ and $V\in\mathbb{R}^{|\qset|\times d}$ are the historical-user and item factor matrices:
\[
\min_{U,V}\ \sum_{(h,q)\in\Omega}(R_{h,q}-U_h^\top V_q)^2
+\lambda_U\lVert U\rVert_F^2+\lambda_V\lVert V\rVert_F^2.
\tag{3}
\]
The regularization weights $\lambda_U$ and $\lambda_V$ penalize large historical-user and item factors, respectively. The latent dimension $d$ is chosen during historical preprocessing and controls the capacity of these representations. Each row $V_q$ represents an item and each row $U_h$ represents a historical user. Clustering the item rows $\{V_q\}$ partitions the catalog into $A$ item arms $\aset=\{1,\ldots,A\}$, with arm $a$ containing $\qset_a$. Clustering historical-user rows, together with metadata where available, yields $C$ latent user groups $\cset=\{1,\ldots,C\}$. For the deployment, $A$ and $C$ were selected before enrollment; Section~\ref{app:field-history} gives those values and the candidate ranges.

For a new user, the learner observes metadata group $g_u$ but not the latent group to which the user belongs. Earlier cohorts estimate the association between metadata and latent groups. The vector $p^{(0)}_u\in[0,1]^C$ therefore has entries $p^{(0)}_{u,c}=\widehat{\Pr}(c\mid g_u)$ and sums to one; an unseen metadata value receives the global group frequencies. For each group--arm pair, the historical success and failure counts $s^{\mathrm{hist}}_{c,a}$ and $f^{\mathrm{hist}}_{c,a}$ give a smoothed mean and a fixed-strength Beta prior. Here, $\alpha_0$ and $\beta_0$ are smoothing pseudo-counts, and $\kappa>0$ is the selected effective prior strength:
\[
\widehat\mu_{c,a}=
\frac{\alpha_0+s^{\mathrm{hist}}_{c,a}}
{\alpha_0+\beta_0+s^{\mathrm{hist}}_{c,a}+f^{\mathrm{hist}}_{c,a}},\qquad
(\alpha^{H}_{c,a},\beta^{H}_{c,a})=
\kappa(\widehat\mu_{c,a},1-\widehat\mu_{c,a}).
\tag{4}
\]
The quantities $\alpha^H_{c,a}$ and $\beta^H_{c,a}$ are the historical success and failure pseudo-counts for group $c$ and arm $a$. For a new user, the initial arm belief aggregates uncertain group membership,
\[
\bigl(\alpha^{(0)}_{u,a},\beta^{(0)}_{u,a}\bigr)
=\sum_{c=1}^{C}p^{(0)}_{u,c}\bigl(\alpha^H_{c,a},\beta^H_{c,a}\bigr).
\tag{5}
\]
Equation~(5) summarizes the metadata-conditioned mixture by a single Beta distribution. It preserves the mixture's expected reward for arm $a$, the center of the warm-started Thompson distribution before the user has supplied feedback. Fixing $\kappa$ gives every user--arm prior the same, controlled amount of historical influence, independent of how many observations happened to be recorded for a historical group--arm cell. This first-moment match gives comparable initialization across users while retaining one Beta draw per arm for slate construction. We examine the effect of the chosen prior strength in the transfer sensitivity analysis.

\subsection{Constraint-aware slate construction}
\label{sec:slates}

Because an item can be displayed only once to a user, a promising arm's finite per-user inventory can be exhausted early. Given the campaign horizon, the learner therefore paces each arm's consumption while constructing a slate, balancing sampled reward now against feasible items in later rounds. Thompson sampling makes this reward--information trade-off explicit: a Beta belief represents uncertainty about an arm's binary reward rate, and an independent draw from each arm's belief is a plausible reward rate for the next display. A high draw favors the arm, while uncertain arms occasionally draw high enough to be explored. At selection step $i\in[K]$, let $\qset^{\mathrm{avail}}_{u,t,i}$ be the items that user $u$ has not seen and that have not yet been selected for the current slate. Let $m_a=|\qset_a|$ be the per-user inventory of arm $a$, let $h_{u,a,t,i}$ be the number of earlier selections in the current slate that used arm $a$, and define
\[
d_{u,a,t,i}=1-\frac{|\qset_a\cap\qset^{\mathrm{avail}}_{u,t,i}|}{m_a},
\qquad r_{u,a,t,i}=1-d_{u,a,t,i}.
\]
Thus, $d_{u,a,t,i}$ is the share of user $u$'s arm-$a$ inventory already spent and $r_{u,a,t,i}$ is its remaining share before selection step $i$. An arm is feasible exactly when $\qset_a\cap\qset^{\mathrm{avail}}_{u,t,i}\neq\varnothing$. To pace inventory use, the learner carries a smoothed arm-specific pacing error $\bar e_{u,a,t-1}$ from the preceding round, initialized to zero: $e_{u,a,t,i}=\rho\bar e_{u,a,t-1}+(1-\rho)(d_{u,a,t,i}-t/T)$. Here $\rho\in[0,1]$ retains prior error; $\operatorname{clip}(x,l,h)=\min\{\max\{x,l\},h\}$ limits the adjustment to bounds $\eta_{\min}$ and $\eta_{\max}$; and $\phi\geq0$ sets its strength. For every feasible arm, the policy samples $\widetilde\theta_{u,a}\sim\mathrm{Beta}(\alpha_{u,a},\beta_{u,a})$ and computes
\[
\begin{aligned}
D_{u,a,t,i}&=1+\phi\,\operatorname{clip}(e_{u,a,t,i},\eta_{\min},\eta_{\max}),\quad
\operatorname{score}_{u,a,t,i}=\widetilde\theta_{u,a}/P_{u,a,t,i},\\[-2pt]
P_{u,a,t,i}&=1+\gamma h_{u,a,t,i}+\delta(1-r_{u,a,t,i})D_{u,a,t,i}.
\end{aligned}
\tag{6}
\]
Here, $D_{u,a,t,i}$ is the pacing multiplier and $P_{u,a,t,i}$ is the total penalty that divides the Thompson draw. The coefficient $\gamma\geq0$ discourages repeated arm use within one slate, while $\delta\geq0$ weights the depletion penalty. The controller compares consumption with the uniform plan $t/T$; a positive error means an arm is being used faster than plan and strengthens its depletion term. After the slate, it stores $\bar e_{u,a,t}=\rho\bar e_{u,a,t-1}+(1-\rho)(d_{u,a,t,K+1}-t/T)$. The score therefore trades immediate sampled reward against within-slate variety and later feasibility.

After selecting the highest-scoring arm, the policy samples one item uniformly from that arm's available pool and removes it before the next selection step. Uniform sampling treats items within an arm as exchangeable. An arm may appear more than once when it still has unseen items, which is necessary when $K>A$.

\begin{algorithm}[t]
\caption{Constraint-aware slate construction for user $u$}
\label{alg:slate}
\begin{algorithmic}[1]
\Require Current Beta parameters, unseen arm pools, slate size $K$
\State $S\gets\varnothing$ and $\mathcal V\gets\qset\setminus\qset^{\mathrm{seen}}_{u,t-1}$
\For{$i=1$ to $K$}
  \State $F\gets\{a:\qset_a\cap\mathcal V\neq\varnothing\}$
  \For{each $a\in F$}
    \State Sample $\widetilde\theta_{u,a}$ and compute Equation~(6)
  \EndFor
  \State $a^*\gets\arg\max_{a\in F}\operatorname{score}_{u,a,t,i}$
  \State Sample $q$ uniformly from $\qset_{a^*}\cap\mathcal V$
  \State Set $S\gets S\cup\{q\}$, set $\mathcal V\gets\mathcal V\setminus\{q\}$, and update the within-slate state
\EndFor
\State \Return $S$
\end{algorithmic}
\end{algorithm}

\subsection{Posterior adaptation across rounds}
\label{sec:updates}

Recommendations are commonly committed as a full slate before any item in that slate has been answered, and operational workflows can prepare slates periodically from one parameter snapshot. We therefore hold the model fixed while a round is served and update it only at scheduled feedback checkpoints. All active users receive their $K$-item slates from the same snapshot, so feedback from one item cannot change another item in that slate or another user's slate in the same round. When the round closes, the learner aggregates observed outcomes, refreshes memberships and pseudo-counts, and uses the resulting snapshot for later slates. In the deployment, a round was a daily quiz session: answers from a completed active day were batched at the daily boundary, and membership refresh began after a user had answered at least 10 items. For user $u$ and arm $a$, $s^{(t)}_{u,a}$ and $f^{(t)}_{u,a}$ are the scalar counts of correct and incorrect outcomes in round $t$. Their cumulative values are $s^{(\leq t)}_{u,a}=\sum_{\tau\leq t}s^{(\tau)}_{u,a}$ and $f^{(\leq t)}_{u,a}=\sum_{\tau\leq t}f^{(\tau)}_{u,a}$.

At each such feedback checkpoint, the learner computes the membership-probability vector $p^{(t)}_u\in[0,1]^C$ from the registration-time mixture and cumulative evidence:
\[
\begin{aligned}
\ell^{(t)}_{u,c}&=\log p^{(0)}_{u,c}
+\sum_a\log B\!\left(s^{(\leq t)}_{u,a}+\alpha^H_{c,a},
f^{(\leq t)}_{u,a}+\beta^H_{c,a}\right)\\[-2pt]
&\quad-\sum_a\log B(\alpha^H_{c,a},\beta^H_{c,a}),\\
p^{(t)}_{u,c}&=\operatorname{softmax}_c\!\left(\ell^{(t)}_{u,c}\right).
\end{aligned}
\tag{8}
\]
Here, $\ell^{(t)}_u\in\mathbb{R}^{C}$ is the vector of log-membership scores and $B$ is the Beta function. Under the arm-level conditional-exchangeability model, the Beta-function difference is the log marginal likelihood of a user's group--arm counts after integrating a Bernoulli rate under the historical Beta prior; the binomial coefficient cancels across groups. Thus, $p^{(t)}_u$ is a soft reassignment under a working arm-level model. The entries of $p^{(t)}_u$ sum to one. Between checkpoints, the learner uses the most recently computed membership vector. It also maintains two nonnegative $C\times A$ matrices, $\xi^{+,(t)}$ and $\xi^{-,(t)}$, for cumulative fractional correct and incorrect evidence assigned to each group--arm pair. The shared-update weight $\lambda\in[0,1]$ determines the fraction of a round's evidence placed in these group-level ledgers. The ledgers evolve as
\[
\xi^{\pm,(0)}_{c,a}=0,\qquad
\xi^{\pm,(t)}_{c,a}=\xi^{\pm,(t-1)}_{c,a}+
\lambda\sum_{u\in\uset_t}p^{(t)}_{u,c}x^{\pm,(t)}_{u,a},
\tag{9}
\]
where $x^{+,(t)}_{u,a}=s^{(t)}_{u,a}$ and $x^{-,(t)}_{u,a}=f^{(t)}_{u,a}$.
For each user, the learner also retains that user's fractional contributions $\zeta^{+,(t)}_{u,c,a}$ and $\zeta^{-,(t)}_{u,c,a}$ to the two ledgers, initialized to zero and incremented by $\lambda p^{(t)}_{u,c}s^{(t)}_{u,a}$ and $\lambda p^{(t)}_{u,c}f^{(t)}_{u,a}$, respectively. This produces a leave-one-user-out group mean for the next round:
\[
\begin{aligned}
\bar\mu^{(t)}_{u,c,a}&=
\frac{\alpha^H_{c,a}+\xi^{+,(t)}_{c,a}-\zeta^{+,(t)}_{u,c,a}}
{\alpha^H_{c,a}+\beta^H_{c,a}+\xi^{+,(t)}_{c,a}+\xi^{-,(t)}_{c,a}-\zeta^{+,(t)}_{u,c,a}-\zeta^{-,(t)}_{u,c,a}},\\
\bar\mu^{(t)}_{u,a}&=\sum_c p^{(t)}_{u,c}\bar\mu^{(t)}_{u,c,a},\\
\alpha^{(t)}_{u,a}&=\kappa\bar\mu^{(t)}_{u,a}+s^{(\leq t)}_{u,a},\qquad
\beta^{(t)}_{u,a}=\kappa(1-\bar\mu^{(t)}_{u,a})+f^{(\leq t)}_{u,a}.
\end{aligned}
\tag{10}
\]
Thus, scheduled feedback updates the center of a fixed-strength group-informed prior while each user's cumulative outcomes enter its Beta parameters directly. Equations~(8)--(10) define the scheduled soft-mixture update.

\begin{algorithm}[t]
\caption{Cross-cohort \method{} workflow}
\label{alg:workflow}
\begin{algorithmic}[1]
\Require Historical logs $\mathcal{D}^{\mathrm{hist}}$, current cohort metadata
\State Fit Equation~(3); cluster item and historical-user representations
\State Estimate $p^{(0)}_{u,c}$ and group--arm priors with Equations~(4)--(5)
\For{each round $t$}
  \For{each active user $u$}
    \State Serve the slate returned by Algorithm~\ref{alg:slate}
  \EndFor
  \State Collect delayed feedback after the round
  \State Update memberships, shared pseudo-counts, and user beliefs with Equations~(8)--(10)
\EndFor
\end{algorithmic}
\end{algorithm}

\section{Experimental Setup}
\label{sec:experiments}

\subsection{Baselines}
\label{sec:baselines}

Within each generated environment, policies use the same catalog, slate size, user-level no-repeat rule, and delayed-feedback schedule. At each selection step, a policy chooses an arm and samples an unseen item from that arm.

\noindent\textbf{Full \method{}.} The full policy uses a metadata-conditioned soft mixture of historical group priors, refreshes memberships from each user's cumulative feedback, adds a fraction $\lambda$ of feedback to shared group-level ledgers, and applies slate controls when they are studied. Every non-random policy also updates its own user--arm belief from that user's feedback. It is the deployment treatment.
\paragraph{\textbf{Cold-start TS}.} Cold-start Thompson sampling initializes every user--arm belief as $\mathrm{Beta}(1,1)$ and updates it from that user's outcomes.
\paragraph{\textbf{Metadata-only contextual baselines}.} \textbf{Metadata LinUCB} and \textbf{Metadata LinTS} use the one-hot interaction $x_{u,a}=e_{g_u}\otimes e_a$: 20 features in transfer and 136 in the year-split benchmark. Both receive the same source estimates and metadata association as \method{}, collapsed to a metadata--arm mean at $\kappa=10$, and pool feedback by metadata--arm cell after each session. They have no user, latent-group, item, or factor features. LinUCB uses $\alpha=.5$; LinTS uses Gaussian scale $\nu=.5$. They appear only in transfer and the year-split benchmark.
\paragraph{\textbf{Transfer ablations}.} \textbf{Warm TS (fixed mixture)} retains $p_u^{(0)}$ and updates individual arm beliefs, but has neither reassignment nor a group ledger. \textbf{Hard-cluster TS} uses fixed $\arg\max_c p^{(0)}_{u,c}$ assignments; its semi-synthetic counterpart is \textbf{Hard membership}. \textbf{Static source mixture} ranks the initial metadata-conditioned historical mean and never updates.
\paragraph{\textbf{Semi-synthetic ablations}.} \textbf{Global prior} replaces metadata-conditioned frequencies with population frequencies. In the real-catalog scarcity diagnostic, \textbf{No depletion} sets $\delta=0$ while retaining the remaining personalization components. \textbf{Random recommendation} samples uniformly from unseen items and is the deployed control.
\paragraph{\textbf{Slate-control variants}.} The finite-inventory environment fixes highly concentrated, correct beliefs and compares only slate construction. \textbf{No controls} sets all terms to zero; \textbf{Diversity only} sets $\gamma=.5$; \textbf{Depletion only} sets $\delta=\phi=1$; \textbf{Full selector} uses both; and \textbf{No adaptive controller} changes only $\phi$ to zero. Full selector denotes the slate-control rule, not the full cross-cohort policy.
\paragraph{\textbf{Feasible oracle}.} In the transfer and year-split semi-synthetic environments, an oracle knows the simulated reward means while respecting each user's item inventory. It provides the pseudo-regret reference rather than a deployable policy.

\subsection{Three environments}
\label{sec:synthetic}

\begin{figure*}[!t]
\centering
\includegraphics[width=\textwidth]{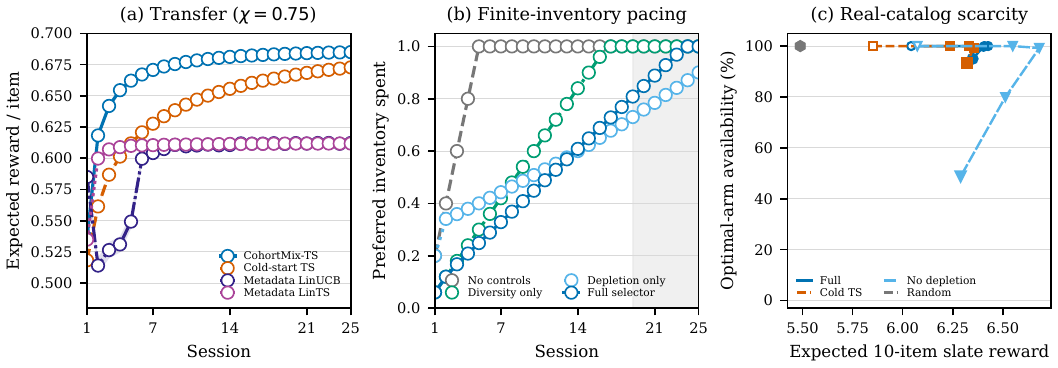}
\caption{One diagnostic for each generated environment. (a) Transfer expected reward at alignment $\chi=0.75$ for \method{}, Cold-start TS, Metadata LinUCB, and Metadata LinTS; higher $\chi$ means closer agreement between historical and target preferences. (b) Preferred-arm inventory use when that arm contains $20\%$ of the $TK$ item opportunities for No controls, Diversity only, Depletion only, and Full selector. (c) Real-catalog scarcity stress in the 2024--2025 year-split environment. Each line joins five consecutive five-session blocks. Both coordinates average over users whose generated optimal arm is the actual 98-item arm: expected 10-item slate reward (horizontal) and the percentage with an unseen item remaining in that arm before the session slate is selected (vertical). Open markers denote sessions 1--5 and filled terminal markers denote sessions 21--25. Shaded bands in (a)--(b) are percentile bootstrap 95\% intervals across 160 cohorts; (c) reports cohort-average trajectories across 96 cohorts.}
\Description{Three panels. The first plots expected reward over 25 sessions for full CohortMix-TS, Cold-start TS, Metadata LinUCB, and Metadata LinTS. The full CohortMix-TS line is highest by the end. The second plots preferred-arm inventory consumed over 25 sessions for four slate-control variants; no controls and diversity only exhaust the arm early, while depletion-aware variants use it gradually. The third plots five connected positions for full CohortMix-TS, Cold-start TS, No depletion, and Random in a real-catalog scarcity stress. Its horizontal axis is expected reward of a ten-item slate for users whose generated optimal arm is a 98-item arm. Its vertical axis is the percentage of those users with an unseen item remaining in that arm before their session slate is selected. An open marker is the first block and a larger filled terminal marker is the fifth.}
\label{fig:generated-environments}
\end{figure*}

The environments answer different parts of the problem while keeping the interaction protocol fixed: users receive 10 unseen items in each of 25 sessions, feedback arrives after the session, and policies are compared on matched generated cohorts. The first asks whether prior-cohort information helps cold-start personalization; the second asks how to use a scarce per-user inventory; and the third asks whether the transfer result persists under a real catalog and a year-to-year split.

\paragraph{Parametric transfer environment.} This is a clean test of historical transfer rather than inventory management. Each observed metadata value contains two hidden user types with opposing preferred arms. Thus, metadata supplies a useful starting point but cannot determine one person's best arm. Every arm contains more items than a user can receive in 25 sessions, and slate controls are inactive. We vary the alignment $\chi$ between historical and target preference profiles: $\chi=1$ means that the prior cohort has the same profile, whereas $\chi=0$ means that it carries no directional information about the target profile. We compare \method{} with the metadata-only contextual baselines, Warm TS (fixed mixture), Hard-cluster TS, Cold-start TS, and Static source mixture. Appendix~\ref{app:synthetic-details} gives the generator, potential-outcome protocol, and prior-strength interaction.

\paragraph{Parametric constraint environment.}
This is a clean test of slate controls rather than a test of learning from history. All policies begin with accurate arm beliefs, but one high-reward arm has a limited number of unseen items for each user. Repeatedly choosing that arm within one session also lowers the reward of later selections in that session. We vary its inventory and compare No controls, Diversity only, Depletion only, Full selector, and No adaptive controller. The resulting trade-off is between taking the scarce arm immediately, spreading choices within a slate, and retaining useful items for later sessions.

\paragraph{Year-split data-calibrated semi-synthetic environment.}
The third environment joins the previous two questions to observed cohort data without replaying any participant's actual recommendation history. A disjoint 2024 cohort supplies the historical factorization, groups, metadata association, and arm priors. A 2025 cohort calibrates hidden target preferences for freshly generated users. Both cohorts use the same 1,347-item catalog, partitioned into eight arms of sizes 128, 239, 137, 84, 298, 98, 243, and 120. The policy sees only 2024-derived structure, a new user's faculty, and simulated feedback; the evaluator alone uses the 2025 records. Each main cohort has 288 users balanced across three source-only preference types, so a majority type cannot mask performance for the others. The main benchmark turns slate penalties off to focus on transfer. Figure~\ref{fig:generated-environments}(c) separately uses the real 98-item arm to study depletion. It compares \method{}, the metadata-only contextual baselines, the listed ablations, and Random. Appendix~\ref{app:synthetic-details} gives the calibration, potential-outcome protocol, and sensitivities.

\subsection{Randomized in-the-wild deployment}
\label{sec:field-setup}

To test whether personalized allocation can match questions to participants, the completed in-the-wild deployment used a 25-day Campus Games quiz. Correctness was the binary reward: questions differ in difficulty and subject familiarity, so a question that suits one participant or group may be a poor match for another. The daily quiz delivered $K=10$ multiple-choice items from a catalog of $Q=1{,}347$ items, with a 20-second limit per item. A prior cohort of 859 users over the same catalog supplied the transferred structure. Historical model selection yielded five item arms and 23 user groups. The five arm sizes were 273, 190, 516, 76, and 292 items. This randomized study evaluates the complete deployed policy against random allocation, rather than an isolated method component.

The previous cohort's questions were sampled independently of recommendations, so no exposure correction is required. For adaptive historical logs with known display propensities, the factorization and group--arm rates can instead use inverse-propensity weighting \cite{liUnbiasedOfflineEvaluation2011}, avoiding a policy-induced exposure pattern being treated as preference.

Assignment occurred once at registration. The 356 treatment users received \method{} slates intended to increase their estimated probability of answering correctly, while 357 control users received random unseen items; quiz length, time limit, scoring, and catalog were identical between arms. Recommendations were planned on a rolling three-day horizon, with assigned items reserved until used or returned to the eligible pool. The deployed settings were $\gamma=0.5$, $\delta=1.0$, $\phi=1.0$, $\eta_{\min}=-0.3$, $\eta_{\max}=0.7$, $\rho=0.3$, $\lambda=0.3$, and $\kappa=10$. Membership updates began only after a user had supplied at least 10 answers. The deployment did not vary $\kappa$.

\section{Results and Empirical Analysis}
\label{sec:results}

\subsection{Cross-cohort transfer depends on prior alignment}
\label{sec:synthetic-results}

Figure~\ref{fig:generated-environments}(a) and Table~\ref{tab:transfer} answer RQ1 and RQ2. When historical and target preferences align, \method{} has the strongest early and campaign reward and the lowest high-regret tail. The transferred mixture gives it useful structure before a user supplies feedback; Cold-start TS begins with diffuse arm beliefs and must first explore. A fixed mixture or hard assignment transfers some structure, but cannot use feedback to resolve initial latent-type uncertainty. Figure~\ref{fig:generated-environments}(a) exposes a failure mode of standard contextual models based only on observed metadata: Metadata LinUCB and Metadata LinTS pool faculty--arm outcomes, so their curves level off when one faculty contains users with opposing hidden preferences. Figure~\ref{fig:metadata-residuals} shows the corresponding source-data pattern, with group--arm correctness varying within metadata categories. The appendix sensitivity check shows that the mean early-reward benefit grows with historical--target alignment; under severe mismatch, its advantage for high-regret users disappears. Thus, RQ1's benefit depends on reliable cross-cohort structure, and RQ2's soft memberships protect against within-metadata heterogeneity.

\begin{table}[t]
\caption{Parametric transfer results at alignment $\chi=0.75$, averaged over 160 generated cohorts. Rewards are expected per displayed item; Minority is campaign reward for the lower-probability hidden subtype. P90 pseudo-regret is the 90th percentile of per-user pseudo-regret at the final session. Bold marks the best result and underline the second-best result in each column.}
\label{tab:transfer}
\centering
\scriptsize
\setlength{\tabcolsep}{1.7pt}
\begin{tabular}{@{}lrrrr@{}}
\toprule
Policy & Early $\uparrow$ & Campaign $\uparrow$ & Minority $\uparrow$ & $P90$ regret $\downarrow$ \\
\midrule
Full \method{} & \textbf{0.622} & \textbf{0.669} & \textbf{0.653} & \textbf{19.18} \\
Warm TS (fixed mixture) & 0.579 & \underline{0.642} & 0.628 & \underline{24.39} \\
Hard-cluster TS & \underline{0.592} & 0.609 & 0.506 & 62.77 \\
Cold-start TS & 0.576 & 0.639 & \underline{0.638} & 24.73 \\
Static source mixture & 0.585 & 0.585 & 0.400 & 87.47 \\
Metadata LinUCB & 0.541 & 0.596 & 0.505 & 65.23 \\
Metadata LinTS & \underline{0.592} & 0.607 & 0.502 & 66.82 \\
\bottomrule
\end{tabular}
\end{table}

\subsection{Slate controls trade early exploitation for later feasibility}
\label{sec:constraint-results}

Figure~\ref{fig:generated-environments}(b) answers RQ3 in the finite-inventory environment. Without depletion control, a policy spends the scarce preferred arm early; variants with the depletion term instead pace it through the final week. The full slate controller accepts less immediate concentration than Diversity only, then has stronger later reward once the scarce arm would otherwise be unavailable. The sensitivity grid in Appendix~\ref{app:synthetic-details} shows that this trade-off weakens when the preferred inventory is less scarce. Depletion is therefore useful for the stated short-inventory setting, rather than a universally reward-maximizing penalty.

\subsection{Year-split data-calibrated semi-synthetic benchmark}
\label{sec:semisynthetic-results}

Table~\ref{tab:semisynthetic} answers RQ1 and RQ2 in the year-split environment. The proposed policy improves on Cold-start TS because the historical mixture starts each user with informative arm preferences, while cold start first explores from a uniform Beta prior. The metadata-only contextual policies improve their faculty-level estimates but cannot separate individual preferences within a faculty, which limits their later gains. Hard membership and a global prior remove, respectively, adaptive uncertainty about type and the metadata signal; both patterns support the value of a soft, metadata-conditioned warm start.

Figure~\ref{fig:generated-environments}(c) answers RQ3 with the actual 98-item arm. The No depletion trajectory gains reward early by repeatedly selecting the preferred arm, then exhausts that arm for many affected users and loses late slate quality. The tempered controller preserves availability into the final block. Random preserves the arm by rarely selecting it, but supplies lower-reward slates throughout. The figure therefore shows the intended early--late pacing trade-off, not a claim that depletion control improves every campaign-wide outcome.

\begin{table}[t]
\caption{Year-split data-calibrated semi-synthetic results over 96 generated cohorts. Source-only 2024 structure initializes each policy; 2025 records calibrate hidden target preferences. Reward entries are expected per displayed item; pseudo-regret is per user over 250 displayed items. The main benchmark turns slate penalties off to isolate transfer; Figure~\ref{fig:generated-environments}(c) reports the separate real-catalog scarcity stress. Bold marks the best result and underline the second-best result in each column.}
\label{tab:semisynthetic}
\centering
\scriptsize
\setlength{\tabcolsep}{2.5pt}
\begin{tabular}{@{}lrrrr@{}}
\toprule
Policy & Early $\uparrow$ & Campaign $\uparrow$ & Late $\uparrow$ & Pseudo-regret $\downarrow$ \\
\midrule
\method{} & \textbf{0.597} & \textbf{0.627} & \textbf{0.632} & \textbf{37.71} \\
Cold-start TS & \underline{0.584} & \underline{0.618} & \underline{0.628} & \underline{39.94} \\
Metadata LinUCB & 0.580 & 0.590 & 0.601 & 46.84 \\
Metadata LinTS & 0.580 & 0.594 & 0.598 & 45.99 \\
Hard membership & 0.583 & 0.615 & 0.625 & 40.59 \\
Global prior & \underline{0.584} & 0.615 & 0.624 & 40.67 \\
Random & 0.565 & 0.565 & 0.565 & 53.12 \\
\bottomrule
\end{tabular}
\end{table}

\begin{figure*}[!t]
\centering
\includegraphics[width=\textwidth]{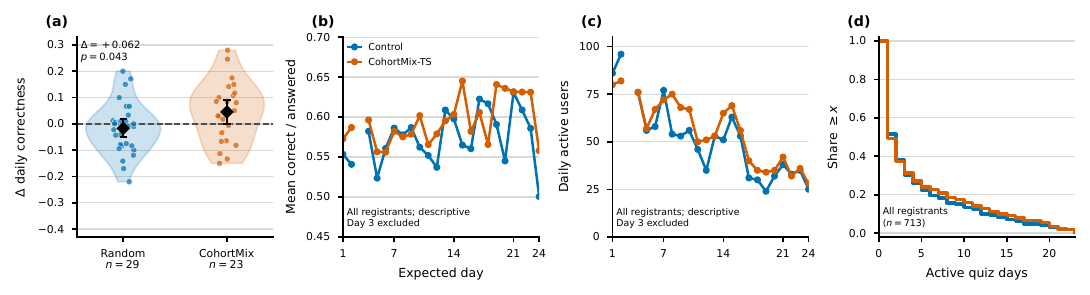}
\caption{In-the-wild deployment outcomes. (a) Participant-level early--late daily-correctness change $\Delta_u$ in the restricted, complete-window analysis ($n=52$; treatment $=23$, control $=29$). Points are users; diamonds and bars are group means and bootstrap 95\% intervals. The treatment--control difference is $0.062$ (95\% interval $[0.005,0.119]$; $p=0.043$). (b) Daily mean correct/answered rate, averaged within each responding user-day; (c) daily active users; and (d) active-day survival. Panels (b)--(d) are descriptive all-registrant traces ($n=713$) after removing technical-failure dates and have changing daily denominators. Blue denotes random control and orange denotes \method{}.}
\Description{Four panels. The first uses violin plots, points, and mean confidence intervals to show early--late daily correctness change for 29 random-control and 23 CohortMix-TS users. The treatment distribution and mean are above the control distribution and mean. The second panel shows descriptive participant-day correct-answer rates, the third the number of daily active users, and the fourth the share of all registrants with at least a given number of active quiz days. Blue represents control and orange represents CohortMix-TS throughout.}
\label{fig:field-deployment}
\end{figure*}

\subsection{Randomized in-the-wild deployment}
\label{sec:field-results}

Real-world participation constraints substantially reduced the size of the correctness analysis. Of the 713 registered users, 385 interacted with at least one quiz item. The outcome analysis excluded activity before the official start and two dates affected by technical failures, then retained participants with at least 30 answered items. This answer-count filter produced an analysis set of 211 users: 109 treatment and 102 control. Because the filter occurs after assignment, the resulting comparison is not an intention-to-treat estimate for all registrants.

The four largest metadata strata contained 58 treatment and 61 control users. Computing $\Delta_u$ additionally requires observed outcomes in both the early and late windows; only 23 treatment and 29 control users met this requirement. Mean $\Delta_u$ was $0.0455$ for treatment and $-0.0168$ for control, a difference of $0.0623$ (bootstrap 95\% interval $[0.005,0.119]$). A two-sided Mann--Whitney rank-sum test returned $p=0.043$. Figure~\ref{fig:field-deployment} shows this distribution and descriptive daily context. Broader responder analyses weakened the evidence: the all-faculty answer-qualified set ($n=93$) had a $0.034$ difference ($p=0.146$), and all complete-window responders ($n=96$) had a $0.033$ difference ($p=0.153$); answer-weighting the restricted outcome gave $0.057$ ($p=0.053$). The small complete-window subset and post-assignment filtering limit precision and interpretation; the result is sensitive to the responder definition and does not isolate individual method components.

We also compared two randomized-arm participation indicators using all 713 registrants. At least one quiz item was started by $53.1\%$ of treatment users (189/356) and $54.9\%$ of control users (196/357; two-sided Fisher test $p=0.652$). At least 30 items were answered by $30.6\%$ of treatment users (109/356) and $28.6\%$ of control users (102/357; $p=0.566$). These are intention-to-treat operational outcomes; neither comparison provided evidence of a treatment difference. Correctness is undefined for users who did not answer, so the study data do not identify an all-randomized correctness effect without a missingness model. We have no pre-assignment quiz outcome.

We do not use the daily count of correct answers as a treatment-effect outcome. Such counts combine answer correctness with changing participation and therefore cannot distinguish improved matching from greater exposure.

\section{Discussion and Limitations}
\label{sec:discussion}

The field study randomized the complete policy against random recommendation, but its correctness result is conditional on post-assignment responses. The all-registrant participation comparisons are intention-to-treat operational outcomes, not correctness or retention effects. Nonresponders have no correctness value to include, and we did not fit a pre-specified missing-data or longitudinal model; we also have no pre-assignment quiz outcome. The restricted result is therefore compatible with improved item matching over time, but not with a population-wide causal claim about learning, interest, satisfaction, or retention.

The $\kappa$ sweeps show that a fixed concentration must be calibrated to transfer alignment: stronger priors help under aligned transfer but can harm under severe source--target mismatch. The finite-inventory isolation deliberately removes prior uncertainty, and the deployment uses one fixed value. Equation~(5) is not a second-moment match, and we did not fit $\kappa$ by empirical Bayes, as in empirical-Bayes meta-prior learning \cite{nabiMetaPrior2022}. The uniform $t/T$ plan is the only pacing target evaluated: we did not test front-loaded, concave, value-aware, attendance-aware, or annealed schedules. The real-catalog stress shows a late-availability benefit from depletion, but it does not establish a general choice of controller weight. Likewise, no difficulty-bucket or item-level posterior was compared with uniform within-arm sampling.

The comparison set is mechanism-specific rather than a broad recommender benchmark. It includes Metadata LinUCB and Metadata LinTS in the transfer and year-split semi-synthetic benchmarks, initialized from the same metadata--arm source estimates and evaluated with the same feasibility and slate constraints. These policies pool feedback within metadata--arm cells and therefore do not represent a user-specific latent state. We do not compare against contextual models with common user--item feature representations or against hierarchical Bayesian TS; \textbf{Hard-cluster TS} is an ablation, not a substitute. The arm partition is fixed throughout the study; handling catalog changes requires a principled transfer of priors across revised arms. Future work should compare richer models under common historical inputs and pre-specify longitudinal field outcomes.

\section{Conclusion}
\label{sec:conclusion}

We developed \method{} for short-horizon, finite-inventory cohort recommendation, transferred historical structure into soft priors, and deployed the complete policy in a randomized campaign. The controlled environments and 2024--2025 data-calibrated benchmark identify settings in which historical warm starts improve early recommendation and a scarce arm benefits from pacing. In our in-the-wild deployment, we observed a larger early--late correctness change only in the reported restricted subset. Future work will test richer representations, adaptive priors and pacing, and longer-running services with pre-specified longitudinal outcomes.

\appendix
\section{Detailed Environments and Ablations}
\label{app:synthetic-details}

\paragraph{Common protocol.}
All policies construct every user's slate from the same pre-session state and receive feedback only after that session is complete. Within each generated cohort, policies share the same user metadata, item pools, and Bernoulli potential outcomes. We summarize uncertainty over independently generated cohorts; ``early'' denotes sessions 1--5 and ``late'' denotes sessions 19--25.

\paragraph{Ablations.}
Warm TS (fixed mixture) keeps the registration-time metadata mixture and updates only a user's local arm counts. \textbf{Hard membership} replaces that mixture with its most likely historical group, while \textbf{Static source mixture} serves the highest-ranked source arm without feedback updates. \textbf{Global prior} replaces $\Pr(c\mid g_u)$ with population group frequencies. \textbf{Metadata LinUCB} and \textbf{Metadata LinTS} pool one-hot metadata--arm estimates and do not maintain a user-specific latent state. In the scarcity diagnostic, \textbf{No depletion} changes only $\delta$ from $.25$ to zero; \textbf{Random} samples feasible unseen items uniformly.

\paragraph{Transfer environment and membership ablations.}
The transfer environment has four metadata values, eight hidden groups, and five arms. Each metadata value places prior mass $.65$ and $.35$ on two groups with opposing preferred arms, so metadata alone cannot identify a user's best arm. Target group profiles interpolate between the historical profile and an orthogonal profile using $\chi$: $\chi=1$ denotes aligned source and target profiles, whereas $\chi=0$ removes this alignment. We draw 400 historical binary outcomes per group--arm cell, generate 480 target users, and give each user 300 items per arm; the catalog is deliberately generous, so slate controls are inactive. We use $T=25$ sessions and $K=10$ items per session. The comparison includes fixed-mixture, hard-membership, static-source, cold-start, and metadata-only contextual alternatives. The $\kappa\times\chi$ grid in Table~\ref{tab:appendix-ablations} shows the early-reward contrast between \method{} and Cold-start TS.

\paragraph{Finite-inventory ablations.}
The finite-inventory environment has 256 users, one preferred arm with mean reward $.70$, and four fallback arms with mean reward $.50$. The preferred arm has capacity $m\in\{25,50,75,100\}$, while each fallback arm has capacity 125. Thus $m/(TK)\in\{.10,.20,.30,.40\}$ is the fraction of a user's campaign slots that the preferred arm could fill. To make slate diversity relevant, every earlier use of an arm within the same slate lowers the reward logit by $.25$. We vary diversity $\gamma\in\{0,.25,.5,1\}$, depletion weight $\delta\in\{0,.5,1,2\}$, and controller strength $\phi\in\{0,.5,1,2\}$. Table~\ref{tab:appendix-ablations} reports the component ablation at $m/(TK)=.20$. The $\delta$ grid shows that stronger depletion control is needed when the preferred arm supplies only $.10$ of the campaign: early exhaustion falls from 1.00 at $\delta\leq1$ to .27 at $\delta=2$, while at $.20$ it is zero for $\delta\geq.5$. At looser capacities, the diversity-only policy can have higher campaign reward, which is why the controller is presented as a pacing trade-off rather than a universal improvement.

\begin{table*}[t]
\caption{Ablation and sensitivity details. (Left) finite-inventory component ablation at $m/(TK)=.20$; rewards are expected per displayed item and Early exhaust is the fraction of users whose preferred arm is exhausted before the final week. ``No adaptive controller'' retains $\gamma=.5$ and $\delta=1$ but sets $\phi=0$. (Right) the early-reward difference between \method{} and Cold-start TS in the transfer environment; positive values favor \method{}. Bold marks the best and underline the second-best result within each left-hand reward column.}
\label{tab:appendix-ablations}
\centering
\scriptsize
\begin{minipage}[t]{0.58\textwidth}
\centering
\begin{tabular}{@{}lrrrr@{}}
\toprule
Policy & Early $\uparrow$ & Campaign $\uparrow$ & Late $\uparrow$ & Early exhaust $\downarrow$\\
\midrule
No controls & .4386 & .4333 & .4320 & 1.00 \\
Diversity only & \textbf{.5244} & \underline{.5002} & .4506 & 1.00 \\
Depletion only & .4756 & .4790 & .4834 & \textbf{.00} \\
No adaptive controller & \underline{.5177} & \underline{.5081} & \underline{.4983} & \textbf{.00} \\
Full selector & .5168 & \textbf{.5083} & \textbf{.4996} & \textbf{.00} \\
\bottomrule
\end{tabular}
\end{minipage}\hfill
\begin{minipage}[t]{0.36\textwidth}
\centering
\begin{tabular}{@{}lrrr@{}}
\toprule
 & $\chi=0$ & $\chi=.5$ & $\chi=1$\\
\midrule
$\kappa=1$  & .033 & .029 & .034\\
$\kappa=5$  & .032 & .034 & .073\\
$\kappa=10$ & .029 & .035 & .083\\
$\kappa=20$ & .025 & .035 & .090\\
$\kappa=40$ & .015 & .032 & .098\\
\bottomrule
\end{tabular}
\end{minipage}
\end{table*}

\paragraph{Sensitivity patterns.}
At $\chi=0$, the warm start still improves average early reward, but it has a worse 90th-percentile user pseudo-regret tail; at $\kappa=10$, the Full-minus-Cold P90 difference is $+12.27$. At $\chi=1$, the corresponding difference is $-19.62$. Thus, alignment chiefly determines whether historical structure also protects users with high regret. In the high-alignment year-split environment, a separate 32-cohort prior sweep over $\kappa\in\{2,5,10,20\}$ improved campaign reward as prior strength increased. A $\lambda\in\{0,.1,.3,.5\}$ check was nearly flat, so it does not identify $.3$ as a uniquely preferred sharing weight. These are targeted sensitivity checks, not a search over all configurations.

\paragraph{Year-split data-calibrated semi-synthetic environment.}
The disjoint 2024 source contains 49,675 answers from 561 users, and the 2025 target calibration contains 23,838 answers from 347 different users. We first shrink source user--arm response profiles toward 2024 arm means and cluster those eight-dimensional profiles into three source-only groups. These groups determine the historical Beta priors and $\widehat{\Pr}(c\mid\text{faculty})$. The 2025 records are used only by the generator to obtain hidden target profiles, user residuals, and target group--arm means; policies receive neither these records nor the target labels. Source-calibrated question effects and sampled target user residuals create fresh potential outcomes, shared across policies. The source--target group--arm correlation is .985. Each cohort has 288 users balanced across hidden types and draws from the actual 1,347-question, eight-arm catalog without replacement for 25 sessions of 10 items. The transfer benchmark fixes $\kappa=10$ and $\lambda=.3$ with slate penalties inactive to isolate transfer. The scarcity diagnostic uses the real 98-item arm and compares $(\gamma,\delta,\phi)=(0,.25,1)$ with $\delta=0$; it evaluates availability only among users for whom that arm is hidden-optimal.

\paragraph{Field deployment details.}\label{app:deployment}\label{app:field-history}

The deployment was a 25-day daily quiz within a Campus Games quiz. On an active day, a participant received 10 previously unseen multiple-choice items from a 1,347-item catalog, with a 20-second limit; incorrect or expired answers received reward zero. The preceding 859-user cohort used the same catalog. Historical preprocessing selected five item arms and 23 user groups from candidate counts 2--7 and 2--39. Their (size, historical correctness) pairs were $(273,.555)$, $(190,.560)$, $(516,.500)$, $(76,.524)$, and $(292,.570)$. Faculty was the deployment-time metadata field; 17 categories occurred in the current cohort.

Randomization occurred once at registration. Control users received uniformly sampled unseen items; treatment users received \method{} slates. Recommendations were planned over a rolling three-day horizon: assigned items were reserved, excluded from later candidate pools, and released if unused. The policy used $\gamma=.5$, $\delta=\phi=1$, $\eta_{\min}=-.3$, $\eta_{\max}=.7$, $\rho=.3$, $\kappa=10$, and $\lambda=.3$; membership updates began after 10 answers.

The de-identified records include event type, time, correctness, and response time; our analysis removed pre-event activity and two unreliable dates. Enrollment records show control/treatment counts of 357/356 registered, 196/189 with any activity, and 102/109 with at least 30 answers. The analysis retained 61/58 users in the four largest metadata strata and 29/23 with both windows. Equation~(2) uses expected days 1--7 and 18--24. All-faculty thresholded responders gave a $0.034$ difference ($p=0.146$); all complete-window responders gave $0.033$ ($p=0.153$). Complete-window response was 13.7\% in control and 13.2\% in treatment ($p=0.913$), but it does not identify missing correctness outcomes.

\begin{figure}[H]
\centering
\includegraphics[width=\columnwidth]{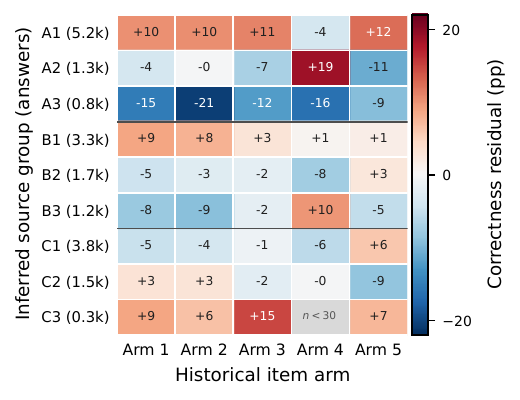}
\caption{Source-only group--arm variation conditional on metadata. Within the three highest-support metadata categories that split into multiple inferred groups, each cell is a group's leave-group-out correctness residual (percentage points) relative to other historical users with the same metadata category and item arm. A--C denote anonymized metadata categories; rows are ordered by source answer support. The gray cell has fewer than 30 observations.}
\Description{A five-column heatmap for historical item arms one through five. Nine rows are grouped into three anonymized metadata categories A, B, and C. Cells show positive or negative leave-group-out correctness residuals in percentage points relative to users with the same metadata category; colors range from blue negative values through white near zero to red positive values. One low-support cell is gray.}
\label{fig:metadata-residuals}
\end{figure}

\paragraph{Historical variation conditional on metadata.}
Figure~\ref{fig:metadata-residuals} gives a descriptive source-only view of group--arm variation after conditioning on the observed metadata category. It is not an independent validation of latent grouping, because the historical grouping procedure also used metadata; the controlled transfer environment isolates that question.

\section*{Ethical Considerations}

The deployment used voluntary participation and a review process appropriate to research involving human participants. The analysis uses de-identified event logs, minimizes retained metadata, and should restrict access to personnel who need the data for the study. Any shared study data should remove identifiers, dates that permit linkage, and small strata that could enable re-identification.

Optimizing correctness can favor easier items for some users and can create unequal experiences if historical metadata encode unequal access, preparation, or participation. The method uses metadata as an uncertain prior rather than a fixed assignment, but this does not remove that risk. Before deployment, operators should define privacy-preserving, adequately sized monitoring groups and report recommendation exposure, correctness, and participation by group alongside aggregate outcomes. We do not make subgroup-effect claims from the deployment: its restricted complete-window analysis contains 52 users across four metadata strata. The system should not use quiz outcomes for grading, discipline, or high-stakes decisions. Operators should provide an explanation of personalization, an opt-out path where feasible, and monitoring for unexpected disparities.

\section*{Acknowledgement}
This work was funded by the Ministerium für Wissenschaft und Gesundheit des Landes Rheinland-
Pfalz through the project AI4Nof1 under grant 724-0019\#2024/0001-150115404.

\bibliographystyle{ACM-Reference-Format}
\bibliography{references,references1}

\end{document}